\documentclass[conference]{IEEEtran}
\IEEEoverridecommandlockouts
\usepackage{cite}
\usepackage{amsmath,amssymb,amsfonts}
\usepackage{algorithmic}
\usepackage{graphicx}
\usepackage{textcomp}
\usepackage{xcolor}
\def\BibTeX{{\rm B\kern-.05em{\sc i\kern-.025em b}\kern-.08em
    T\kern-.1667em\lower.7ex\hbox{E}\kern-.125emX}}
\begin{document}

\title{%
  Uydu Görüntülerinden Yol Bölütleme için Yarı-Gözetimli Alan Adaptasyonu \\
  \huge Semi-Supervised Domain Adaptation for Semantic Segmentation of Roads from Satellite Images}{}

\author{\IEEEauthorblockN{Ahmet Alp Kındıroğlu, Metehan Yalçın, Furkan Burak Bağcı, Mahiye Uluyağmur Öztürk}

\IEEEauthorblockA{\textit\textit{Huawei Turkey R\&D Center} \\
Istanbul, Turkey \\
\{ahmet.alp.kindiroglu, metehan.yalcin, furkan.burak.bagci, mahiye.uluyagmur.ozturk\}@huawei.com
}}

\maketitle

\begin{abstract}
	This paper presents the preliminary findings of a semi-supervised segmentation method for extracting roads from sattelite images. Artificial Neural Networks and image segmentation methods are among the most successful methods for extracting road data from satellite images. However, these models require large amounts of training data from different regions to achieve high accuracy rates. In cases where this data needs to be of more quantity or quality, it is a standard method to train deep neural networks by transferring knowledge from annotated data obtained from different sources. This study proposes a method that performs path segmentation with semi-supervised learning methods. A semi-supervised field adaptation method based on pseudo-labeling and Minimum Class Confusion method has been proposed, and it has been observed to increase performance in targeted datasets.
	
	\textit{Öz}---Bu makale, uydu görüntülerinden yolları çıkarmak için yarı denetimli bir bölütleme yönteminin ön bulgularını sunmaktadır. Yapay Sinir ağları ile imge bölütleme yöntemleri, günümüzde uydu görütnülerinden yolların elde edilmesi için kullanılan en başarılı yöntemlerdendir. Ancak bu modeller, yüksek doğruluk oranları edebilmek için değişik bölgelerden bol miktarda eğitim verisine ihtiyaç duyarlar. Bu verinin yeterli miktarda ya da kalitede olmadığı durumlarda, derin sinir ağları eğitmek için değişik kaynaklardan elde edilen işaretli verilerden öğrenme transferiyle eğitim yapmak genel geçer bir yöntemdir. Bu çalışmada, yarı-gözetimli öğrenme yöntemleriyle yol bölütleme yapan bir yöntem öneriyoruz. Psödo-işaretleme ve En Az Sınıf Karışımı yöntemi tabanlı bir yarı-denetimli alan adaptasyonu yöntemi önerilmiş ve hedeflenen veri kümelerinde başarımı arttırdığı gözlenmiştir.
	
	\textit{Anahtar Sözcükler}---uydu görüntüleri, semantik bölütleme, öğrenme transferi, yarı-gözetimli öğrenme

\end{abstract}


\section{Giriş}
Araç navigasyonu, şehir planlama ve iletişim altyapısı planlama gibi çeşitli uygulamalar, sürekli güncellenen yol veri tabanlarına gereksinim duyarlar. Yüksek çözünürlüklü uydu görüntüleriyle eğitilen yol bölütleme modelleri, bu gereksinimi otomatik bir şekilde ucuz ve hızlı olarak sağlayabilmektedir. 

Yol bölütleme (YB) algoritmaları, uydu resimlerinde yol içeren pikselleri, yol içermeyen arka plan görüntülerinden ayırarak piksel tabanlı sınıflandırma yapmaktadırlar. Yol böltüleme problemi, otoyollar, tali yollar, şehiriçi ve şehirdışı alanlardaki taşıt, yaya ve demir yolları değişik çalışmalarda yol bulma algoritmalarının kapsamına dahil edilmişlerdir. Bu yolların otomatik olarak algılanması, kullanılan algılayıcının tipi, kalitesi ve yakınlık seviyesi ile hedeflenen bölgenin coğrafi özelliklerine göre zorluklar içermektedir. Bu zorluklara örnek olarak yolların uydu görüntülerindeki renk ve dokularının değişkenliği, çevresindeki ağaç ve bina gibi objeler ve bunların gölgeleri, yukarıdan bakınca yola benzeyen yamaç vb gibi çizgiler ve kavşaklar gibi yolların çizgisel yapısının karmaşıklaştığı yapılar gösterilebilir. 

Bu zorlukların üstesinden gelebilecek yol bölütleme modelleri eğitebilmek için yüksek kaliteli işaretler ve veri kümeleri gereklidir. Bu probleme bir çözüm olarak araştırmacılar, Google Maps ve Openstreetmaps gibi kaynaklardan elde edilen işaretlemeleri uydu görüntüleriyle eşleyerek çeşitli veritabanlarında kullanmaktadırlar. Ancak bu tarz yol bölütleme veritabanlarındaki temel bir sorun, işaretlemelerde görülen kabul edilebilir seviyenin üzerindeki hatalardır. Yapılan işaretlemeler ve elde edilen görüntülerdeki tutarsızlıklar, yüksek kaliteli otomatik yol bulma modellerinin eğitilmesindeki bir problemdir. Benzer şekilde elde edilen işaretli kaynakların, her tip kaynak, kamera veya bölge için mevcut olmaması değişik kaynaklardan elde edilen verilerin birlikte kullanıldığı yöntemleri yol bulma modelleri için popüler bir seçenek haline getirmiştir. 

Bu çalışmada sunduğumuz katkılar aşağıdaki şekilde sıralanmaktadır:
\begin{itemize}
\item Hedeflenen bir bölge için geliştirilen baz sınıflandırma modelleri, hedef bölgede ve açık kaynaklı veri kümelerinde başarımlarını en iyilemek için sınanmıştır.
\item Psödo-işaretleme ve En Az Sınıf Karışımı yöntemi tabanlı bir yarı-denetimli alan adaptasyonu yöntemi önerilmiş ve hedeflenen veri kümelerinde başarımı arttırdığı gözlenmiştir.
\item En az sınıf karışımı yönteminin daha önce görülmemiş resimler üzerinde yarattığı negatif transfer, eğitim süresince kullanılan özel bir ağırlıklandırma stratejisiyle en aza indirgenmiştir. 
\end{itemize}

\section{Literatür Özeti}

Bu bölümde, literatürde mevcut yol bölütleme yöntemleri gözden geçirilmektedir. Uydu görüntülerinden yeryüzü kullanım haritaları ve yollar, binalar gibi yapıların tespiti uzun zamandır üzerinde çalışılan bir problemdir. Bu konular üzerine yapılan çalışmaların detaylı özetleri, \cite{fortier1999survey} ve \cite{chen2022road} gibi literatür tarama makalelerinde incelenebilir. Bizim bu makalede üzerine yoğunlaştığımız semantik bölütleme tabanlı yöntemler, 2015 yılında önerilen Tam Evrişimli Sinir Ağları (FCN)\cite{long2015fully} ve U-şekilli sinir ağları (UNet)  \cite{drozdzal2016importance} mimarilerinin derin sinir ağları ile piksel tabanlı bölütme yöntemlerinin popülerleşmesi sonrasında yaygınlaşmıştır. Bu tarz mimarilere sahip derin sinir ağlarının eğitilmesi, PASCAL VOC 2012 \cite{everingham2010pascal}, Cityscapes \cite{cordts2016cityscapes} ve MS COCO \cite{lin2014microsoft} gibi piksel seviyesinde işaretli on binlerce resimli verikümelerinin ortaya çıkmasıyla mümkün olmuştur. Benzer şekilde, uydu görüntülerinden yol tanıma problemi üzerine yapılan derin öğrenme tabanlı çalışmaların yapılabilmesine olanak sağlayan açık kaynaklı yol bölütleme veri kümelerinin bir listesi Tablo \ref{tab:veritabanlari}'de listelenmektedir. 

\begin{table}[]
	\centering
	\caption{}
	\label{tab:veritabanlari}
	\resizebox{\columnwidth}{!}{%
		\begin{tabular}{llccccc}
			Çalışma                      & Verikümesi Adı            & \multicolumn{1}{l}{Sınıf Sayısı} & \multicolumn{1}{l}{Yıl} & \multicolumn{1}{l}{Resim \#} & \multicolumn{1}{l}{Bölge \#} & \multicolumn{1}{l}{Piksel Uzaklığı} \\

			\cite{MnihThesis}            & Massachusets Road  & 2                                & 2020                    & 1157                             & 1                                & 1                                   \\
			\cite{baier2021synthesizing} & Geonrw                    & 11                               & 2020                    & 7782                             & 43                               & 1                                   \\
			\cite{wang2021loveda}        & Loveda                    & 7                                & 2021                    & 2522                             & 8                                & 0.3                                 \\
			\cite{kindiroglu2022yeryuzu} & HW\_Rural                  & 21                               & 2022                    & 1079                             & 4                                & 2.4                                 \\
			\cite{kindiroglu2022yeryuzu} & HW\_Urban                  & 21                               & 2022                    & 2720                             & 4                                & 0.6                                 \\
			\cite{xia_2023_openearthmap} & OpenEarthMap              & 8                                & 2022                    & 4031                             & 97                               & 0.25-0.5                           
		\end{tabular}%
	}
\end{table}

Verilen veritabanları üzerine yapılan çalışmalar incelendiğinde HW\_Rural \cite{kindiroglu2022yeryuzu} veritabanında Deeplab3+ \cite{chen2018encoder} ve Unetplus \cite{zhou2018unet++} yöntemlerinin yeryüzü kullanım haritalrı çıkarmada yüksek başarım elde ettiği görülmektedir. Loveda \cite{wang2021loveda} verikümesinde ise  Unetplus \cite{zhou2018unet++} ve HRNet mimarili semantik bölütleme algoritmalarının en yüksek sonuçları verdiği, değişik çalışmalarda  \cite{wang2021loveda, arnaudo2022hierarchical} beliritilmiştir. 

Farklı kaynakları birleştiren yol bölütleme yöntemlerine örnek olarak, yer seviyesinden görüntülerle uydu görüntülerini birleştiren \cite{Mattyus_2016_CVPR} yöntemi örnek olarak verilebilir. Benzer bir başka çalışmada Mattyus vd. \cite{Mattyus_2015_ICCV} yol bulma için markov random field metodunu kullanmış ve yol, kenar, araba ve yolların devamlılık bilgileriyle bölütleme kalitesini arttırmıştır. \cite{sun2019leveraging} çalışmasında uydu görüntüleri üzerinden yol bölütleme işlemi, gps kitlekaynak gps verilerinden elde edilen işaretlemelerle Unet yapısı kullanılarak yapılmıştır. 

Coğrafi bilgi sistemleriyle kullanılan bazı yöntemlerde, piksel düzeyinde yol bilgisinin topolojik yol yapısı verisine dönüştürülmesi üzerine çalışmalar yapılmıştır. Bu tarz yöntemlerin kullanıldığı çalışmalara örnek olarak pikseller ve kenarlardan doğru parçaları elde edilen klasik yöntemler \cite{fortier1999survey} veya aktif kontörler gibi \cite{liang2019convolutional} yöntemler örnek olarak verilebilir. Yol haritalama üzerine güncel çalışmalara baktığımızda ise, \cite{bastani2018roadtracer,Mattyus_2017_ICCV} gibi yöntemler sinir ağları ile bölütleme sonuçlarını iteratif bir grafik oluşturma algoritmasıyla kullanarak sonuç elde etmektedir. 

İşaretli veri bulmanın zor olduğu veya verilerin resimlerinin ve işaretlerinin bir standarda uymadığı veya dağılımlarının farklı olduğu durumlarda, işaretsiz verilerin kullanılması son zamanlarda popülerleşen bir öğrenme yaklaşımıdır. Yol tanıma için bu tarz yöntemler kullanan çalışmalara Hoyer v.d. \cite{hoyer2022daformer} tarafından geliştirilen DaFormer yöntemi örnek verilebilir. Bu yöntemde az bulunan sınıfların örneklenmesi, ImageNet feature uzaklığı hesaplama ve öğrenme katsayısını yavaşça ısıtma gibi yöntemlerle eğitimi düzenleyerek hedef bölgede başarımı arttırmak için kullanılmıştır. Benzer bir yöntem olan \cite{arnaudo2022hierarchical} makalesinde kaynak ve hedef resimler ve işaretler, sınıf karışımı veri arttırımı yöntemiyle hatayı en aza indirgeyerek arttırılmış, bunun yanında çift başlı bir mimariyle daha detaylı segmentasyon haritaları elde edilebilmiştir. 

\section{Yöntem}

Geliştirilen yöntemi veri elde etme ve ön işleme, yolların anlamsal bölütlenmesi ve değişik kaynak görüntülerden yarı-gözetimli öğrenme olarak üç ana kısımda inceliyoruz. 

\subsection{Verilerin Ön İşlenmesi}

En yüksek başarımı almayı hedeflediğimiz HWLC16 veritabanı dışında, iki farklı açık kaynak veritabanını yol bölütleme eğitim kümemizi iyileştirmek için kullandık. Massachusetts roads verikümesi \cite{MnihThesis} ve Loveda\cite{wang2021loveda} veri kümelerinden elde edilen resimler HWLC16' veritabanıyla aynı çözünürlüğe getirilip ızgara şeklinde örnekler alınarak eğitim ve sınama kümelerine katıldı. HWLC16 ve HWLC18 kümelerindeki piksel seviyesi yol işaretlerinin doğruluğu düşük olduğu için koordinat bilgisi mevcut olan yollar, kalınlığı piksel boyuna orantılı olacak şekilde resim üzerine çizilerek yolların doğruluğu arttırıldı. Eğitim için yapılan ön işleme sırasında her bir resime ait işaretler, yol, yol değil ve işareti belirsiz pikseller olarak 3 sınıfa indirgendi. Bunu takiben [0.5, 1.5] katsayısı ile yeniden boyutlandırılan resimler rastgele olarak döndürme, ters çevirme ve renk değiştirme işlemlerine tabi tutularak 512*512 rastgele kesitler alınıldı.

\subsection{Anlamsal Bölütleme}

Uydu görüntülerinden yol haritaları çıkarımı bir anlamsal bölütleme problemi olarak tanımlanmaktadır. Bu çalışmada, 3 farklı anlamsal bölütleme mimarisi olan Unet++ ve HRNet mimarileri, değişik  kodlayıcılarıla birlikte kullanılarak anlamsal bölütleme yapılmıştır. 

Eğitim sırasında kullanılan yöntemlerden ilki olan Unet++ mimarisi, kodlayıcı ve kod çözücü arasındaki aynı boyuttaki özellik haritalarının(feature map) aktarıldığı atlama yoluna birleşmenin öncesine iç içe yoğun evrişim blokları dizisi eklemiş ve bu atlamaları daha yoğun olacak şekilde güncellemişlerdir. Bu tarz mimariler, genelde öznitelik katmanlarını yüksek çözünürlüklü özniteliklerden düşük çözünürlüklülere doğru oluşturur ve bağlar. HRNet mimarisi \cite{YuanCW19}, paralel bir yapıda dört değişik çözünürlükte öznitelikler oluşturur. Bu mimarideki modeller, değişik çözünürlüklü ve boyutlu şekilleri öğrenmek için çok çözünürlüklü paralel dallar oluşturur ve bunlar arası bilgi alışverişinden yararlanır. Bu özellikle, sınırların ayrıştırılmasının zor olduğu durumlarda yüksek doğrulukla ayrım yapabilecek özniteliklerin çıkarılmasına yardımcı olur. Bu değişik boyuttaki öznitelikler, her aşama sonrasında birbirine bağlanmakta ve en son katmanda tek bir sonuç olarak birleştirilmektedir. 

\subsection{Yarı Gözetimli Öğrenme}

Çalışmalarda yapılan deneylerde, hedeflenen bölgelerde yeterli miktarda yüksek çözünürlüklü verilerin olmaması, bizi daha kaliteli ve daha büyük miktarlarda işaret içeren kaynaklardan veri kullanımına yönlendirmiştir. Değişik kaynaklardan elde edilen verilerin birlikikte öğrenilmesi için en az sınıf karışımı (MCC) \cite{jin2020minimum} yöntemi tabanlu yarı-gözetimli öğretim algoritması kullanılmıştır. 

En Az Sınıf Karışımı yöntemi, kaynak verinin sınıflandırma hatasını, tahminlerin kararlılığını arttırarak en aza indirgemeyi hedefleyen bir yöntemdir. MCC fonksyonu girdi olarak sinir ağları modeli çıktısı olan logit değerlerini alır ve bu fonksyondaki değerleri bir değeri en fazlalayarak diğerlerini bastıracak şekilde değiştirmeyi amaçlar. Bunun için logit matrisini kendisinin tersiyle çarparak sınıf sınıf karışım matrisine yakınsanan bir korelasyon matrisi elde eder. Bu matris üzerinde köşegen değerlerini en fazlalayacak şekilde hesaplanan bir hata fonksyonu, modelin işaretsiz veriler üzerinde sınıf karışımını azalatacak şekilde eğitimini sağlar. Algoritmanın detayları Jin v.d.  \cite{jin2020minimum} makalesinde anlatılmaktadır.

 
Yapılan deneylerde MCC ile eğitimin özellikle öncelikle görülmeyen veriler üzerinde hata fonksyonunun optimal değerlere yakınsanmasını yavaşlattığı gözlemlenmiştir. Bu sebeple kullanılan yöntemde iki iyileştirilme yapılması önerilmiştir. Bunlardan ilki, Denklem \ref{eq:2}'de önerilen alpha değeriyle hata fonksyonunun genel hata fonksyonuna etkisini, iterasyon sayısıyla korelasyon içinde yavaş yavaş arttırılarak kullanılması olmuştur.


\begin{equation}
	L_\varepsilon\bigl(y,f(x,w)\bigr) = L_{SN}\bigl(y_t,f(x_t,w)\bigr) + \alpha * L_{MCC}\bigl(y_s,f(x_s,w)\bigr)  
	\label{eq:2}	
\end{equation}

İkinci olarak kullanılan yöntem ise, uçtan uca eğitim sırasında, işaretsiz verilerin sözde etiketler yaratılarak bu etiketlerin gerçek etiketler gibi sınıf hatası hesaplamada kullanılmasıdır. Eğitim öğrenci-öğretmen yapısı kullanan bir sinir ağı modeliyle yapılır. Öğrenci modelin eğitimş sırasında, işaretli bir veriyle eğitilen öğretmen bir modelden işaretsiz her eğitim resmi için $x^{H,W,N}$ bir etiket tahmin edilir. Bu etiketler, etiketsiz verilerle yarı gözetimli öğrenme fonksyonuna çapraz entropi hatası olarak eklenmektedir.


\section{Verikümeleri ve Deneyler}

Bu bölümde yöntemimizin performansını 3 farklı kaynaktan elde edilen veritabanları üzerinde sınıyoruz. Öncelikle  her veritabanında kendi eğitim kümeleri üzerinde eğitim yapılıp, sınama kümeleri üzerinde taban başarım değerleri incelenmiştir. Bu deneyleri takiben, yarı-gözetimli öğrenme transferi yöntemleri kullanarak modeller genelleştirilmeye çalışılmış ve bu deneylerin sonuçları değişik veri tabanlarına ait sınama kümelerinde gözelnmiştir.

\subsection{Verikümeleri}
\subsubsection{HWLC16 ve HWLC18  Verikümeleri}

Huawei Landcover Level 16 (HWLC16) verikümesi, \cite{kindiroglu2022yeryuzu} çalışmasında verilen ve Çin'in Guangxi bölgesinden elde edilen uydu resimleri ve işaretlemelerini içermektedir. HWLC16 veri tabanı 7 farklı kırsal bölgeden elde edilen piksel başına 2.4 metre çözünürlüklü resimlerden oluşmaktadır. 642 adet 5000*6000 veya daha küçük eğitim ve 110 adet aynı çözünürlükte sınama imgesi içeren veri tabanında işaretler resimlerden daha düşük çözünürlüğe sahiptir. Bu veritabanındaki resimlerden, 640*640 boyutlarında 33895 eğitim ve 9955 sınama resmi çıkarılarak tüm deneylerde kullanılmıştır. HWLC16 veritabanındaki yol işaretlemeleri, diğer verikümelerine kıyasla yüksek hata oranına sahiptir ve kapsanan alanda mevcut tali yolların yarısından fazlasına ait işaretler, veritabanında işaretlenmemiştir.

Aynı kaynaktan elde edilen ikinci bir verikümesi olan Huawei Landcover Level 18 (HWLC18) verikümesi, benzer şekilde 7 farklı şehirden bölgeler içermektedir. Şehir bölgerinde daha yüksek detaylı işaretlere duyulan ihtiyaç yüzünden HWLC16'ya kıyasla daha detaylı ancak daha az alanı kapsayan görsel ve işaretlemelere sahip resimlerden oluşmaktadır. Piksel başına 0.6 metre çözünürlüğü bulunan bu verikümesinde  1079 adet 5000*6000 eğitim ve 121 adet aynı çözünürlükte eğitim resmi mevcuttur. Bu veritabanından elde edilen resimlerden, HWLC16 resimleriyle uyumlu olması amacıyla 4 kat küçültülerek 2720 adet 640*640 eğitim ve 671 adet sınama imgesi elde edilmiştir. Sayıca az olmasına rağmen bu veritabanındaki işaretler, HWLC16 verikümesindekilerin aksine yüksek doğrulukta işaretlenmiştir.

Her iki veritabanında da resimlere ait piksel başı 2.4 metre çözünürlüğe sahip 21 sınıflı semantik bölütleme işaretleri mevcuttur. Bu çalşmada mevcut 21 sınıftan sadece yol sınıfına ait işaretler kullanılmıştır. 

\subsubsection{Loveda Verikümesi}

Loveda verikümesi \cite{wang2021loveda}, Wuhan Üniversitesi tarafından Çin'in kırsal ve şehir bölgelerinden toplanmış açık kaynaklı bir yeryüzü kullanım verikümesidir. HWLC16 ve HWLC18 verikümelerine benzer şekilde Urban ve Rural isimli iki alt kümesi mevcuttur. Verikümesindeki resimler piksel başına 0.3 metre detay seviyesine sahip ve 1024*1024 çözünürlüktedir. Loveda verikümesinde Urban kümesinde 1156, Rural kümesinde ise 1366 resim mevcuttur.

\subsubsection{Massachusets Yol Verikümesi}

Massachusets Yol Verikümesi \cite{MnihThesis}, yukarıda verilen resimlerden farklı olarak Amerika'dan uydu görüntüleri içeren 1 metre çözünürlüklü uydu görüntüleri ve bu görüntülere ait işaretlerden oluşmaktadır. Eğitim ve sınama kümelerinde 1500*1500 çözünürlüklü 1108 ve 49 resim mevcuttur.

\subsection{Uygulama Detayları ve Değerlendirme Kriterleri}

HRNet ve Unet++ mimarilari kullandigimiz deneyler pytorch kutuphanesi ile 11Gb grafik karti hafizali Geforce Nvidia 2080TI kartiyla yapilmistir. Sinir aglarimizi AdamW optimizasyon stratejisi ve 12 mini-batch boyu ile egittik. Modellerimizde öğrenme hızı olarak 0.0001 değerini kullandık ve gelitşirilen her model için 300 bin iterasyon sonunda tüm eğitimlerimizi sonlandırdık. Modellerimizi değerlendirmek için IoU skoru kullandık. Bu değerlendirmelerde sınama aşamasında başarımları arttırmak için herhangi bir sınama veri artırırmı yöntemi kullanmadık.    

\subsection{Deneysel Sonuclar}

Deneylerde kullanılan veritabanları üzerinde yapılan ilk deneyler, bu veritabanlarının birbirinden bağımsız olarak taban sonuçlar bulmayı hedeflemiştir. Her bir veritabanı kendi eğitim kümesi üzerinde eğitilip, eğitim kümesinde bulunmayan resimlerden oluşan kendi sınama kümesi üzerinde sonuçlar elde edilmiştir. Bu Deneylerin sonuçları Tablo \ref{tab:resluts1}'de verilmiştir.

\begin{table}[!htpb]
\centering
\caption{Alan adaptasyonu içermeyen Unet++ ve HRNET modellerinin hedef veritabanları üzerindeki sonuçları}
\label{tab:resluts1}
\resizebox{\columnwidth}{!}{%
\begin{tabular}{|c|c|c|c|c|}
\hline
\textbf{Dataset} & \textbf{Model} & \textbf{Yol IOU} & \textbf{Kırsal IOU} & \textbf{Şehir IOU} \\ \hline
HWLC16             & Unet++         & 17.11             & 17.11              & 6.42               \\ \hline
HWLC16             & HRNet          & 15.77             & 15.77              & 6.56               \\ \hline
HWLC18             & Unet++         & 40.2              & 7.3                & 40.2               \\ \hline
HWLC18             & HRNet          & 44.41             & 8.2                & 44.41              \\ \hline
HWLC1618           & Unet++         & 47.8              & 47.8               & 30.6               \\ \hline
HWLC1618           & HRNet          & 40.5              & 40.5               & 26.3               \\ \hline
Loveda Kırsal     & Unet++         & 47.4              & 32.8               & 16.9               \\ \hline
Loveda Kırsal     & HRNet          & 53.2              & 11.7               & 9                  \\ \hline
Loveda Şehir     & Unet++         & 61.39             & 8.2                & 14.4               \\ \hline
Loveda Şehir     & HRNet          & 62.07             & 8.2                & 15.1               \\ \hline
Mass. Road       & Unet++         & 50.72             & 6.6                & 19.6               \\ \hline
Mass. Road       & HRNet          & 50.17             & 4.43               & 19.2               \\ \hline
\end{tabular}%
}
\end{table}

Deneylerde, her veritabanının kendi üzerindeki başarımının (Yol IoU) yanında HWLC16 (Kırsal IoU) ve HWLC18 (Şehir IoU) sınama kümeleri üzerindeki başarıları da raporlanmıştır. Her verikümesi için hem Unet++ hem de HRNet yapılarında modeller eğitilmiştir. Çalışmada ilk görülen sonuç, HWLC16 veritabanı üzerinde elde edilen başarının diğer veri kümelerine kıyasla düşük olduğudur. Bunun sebebi olarak her ne kadar sayıca en büyük veritabanı olsada bu veritabanındaki işaretlerdeki gürültü ve hata miktarı gösterilebilir. 

Kullanılan HRNET modeli, UNet++ modelinden kırsal bölgelerde daha düşük başarı elde ederken şehir bölgelerinde daha yüksek başarı elde etmiştir. Bu sonuçlardan verilerin gürültüsünün daha yüksek olduğu durumlarda Unet++ modelinin, veride gürültünün daha küçük olduğu durumlarda ise HRNet modelinin daha yüksek performanslı olduğu çıkarımı yapılabilir. 

Yapılan üçüncü bir çıkarım ise, işaretleriyle birlikte HWLC16 ve HWLC18 veri kümelerini beraber kullanmanın, kırsal alanda her iki verikümesini de tek tek kullanmaya kıyasla daha yüksek başarı elde edilebilmesine olanak sağladığıdır. 

Tablo \ref{tab:result2}'de yapılan deneylerde, başarımın en iyilenmesinin hedeflendiği HWLC verikümelerinde yarı-gözetimli öğrenme transferi üzerine yapılan deneylerin sonuçları paylaşılmaktadır.

\begin{table}[!htpb]
	\centering
	\caption{Huawei Veritabanları üzerinde datasetler arası öğrenme transferi sonuçları. Bu deneylerde hedef eğitim kümesi işaretlerle, hedef veri kümesi ise işaretlemeler olmadan kullanılmıştır.}
	\label{tab:result2}
	\resizebox{\columnwidth}{!}{%
		\begin{tabular}{|l|l|c|c|c|c|c|}
			\hline
			Hedef   Eğitim & Kaynak        & Loveda Kırsal & Loveda Şehir & \multicolumn{1}{l|}{HWLC16 } & \multicolumn{1}{l|}{HWLC18} & \multicolumn{1}{l|}{Mass. Yol} \\ \hline
			HWLC16           & -             & 4.1           & 0.1          & 15.8                              & 6.6                              & 0.1                            \\ \hline
			HWLC16           & HWLC18          & 2.9           & 1.2          & 14.8                              & 16.2                             & 6.4                            \\ \hline
			HWLC16           & Loveda Kırsal & 35.6          & 0.0          & 15.1                              & 5.8                              & 0.1                            \\ \hline
			HWLC18           & -             & 9.4           & 21.0         & 7.3                               & 40.2                             & 26.4                           \\ \hline
			HWLC18           & HWLC16          & 20.7          & 23.6         & 30.9                              & 26.4                             & 31.1                           \\ \hline
			HWLC18           & Loveda Şehir  & 0.0           & 33.7         & 7.3                               & 41.1                             & 27.1                           \\ \hline
			HWLC1618         & -             & 41.4          & 45.9         & 40.5                              & 26.3                             & 28.2                           \\ \hline
			HWLC1618         & Loveda Kırsal & 43.2          & 40.0         & 43.3                              & 26.3                             & 18.3                           \\ \hline
			HWLC1618         & Loveda Şehir  & 32.7          & 52.4         & 39.9                              & 38.3                             & 29.0                           \\ \hline
		\end{tabular}%
	}
\end{table}

Tabloda kaynak ve hedef olarak kullanılan bölgelerden HWLC ve Loveda veritabanları Çin'in çeşitli bölgelerinden kırsal ve şehirlere ait alanlardan, Massachustes Yol verikümesi ise bunlara göre daha farklı bölgelerden geliyor. HWLC16 verilerine ait ilk üç satırdaki sonuçlara baktığımızıda bu veritabanındaki işaretlerin kalitesizliği yüzünden hedef eğitim kümesi olarak kullanıldığı durumlarda modelin çok düşük başarım sağladığını gözlemliyoruz. HWLC18i kullanınca başarım \% 15.8'den \% 30.9'a, yeniden boyutlandırılmış HWLC1618'i kullanınca ise IoU \% 40.5'e artıyor. Bunların yanında pseudo işaretleme ve yarı gözetimli öğrenme yöntemleriyle Loveda Kırsal veritabanını eklemek HWLC16 sınama setinde aldığımız en yüksek sonuç olan \% 43.3  IoU'yu bize veriyor. 

HWLC18 veritabanını hedef olarak kullandığımız deneylerde HWLC18 ve loveda şehir kümelerinde yüksek sonuçlar aldığımızı görüyoruz. Bu deneylerde zaten \% 40.2 olan IoU ancak kaynak olarak loveda Şehir kümesini kullandığımız durumda bunu \% 41.1 ile geçiyor. HWLC1618 verisetini hedef eğitim kümesi seçtiğimiz deneyler ise hem Loveda veritabanlarında diğer veritabanlarına göre yüksek sonuçlar vermiştir. 

İkinci olarak,  HWLC verikümelerinin açık kaynak veritabanları üzerinde yarı-gözetimli öğrenme transferi ile kullanımı üzerine çalışmalar gerçekleştirilmiştir. Bu çalışmalar, konum ve kentsel-kırsal alanların dağılımları ve hedeflenen bölge yönünden hedef veritabanlarımıza en çok benzeyen LoveDa  verikümesi üzerinde ve yol bölütleme üzerine pek çok referans çalışmada kullanılan Massachustests Yol verikümesi üzerinde gerçekleştirilmiş ve sonuçları Tablo \ref{tab:result3}'de sunulmuştur.

\begin{table}[!htpb]
	\centering
	\caption{}
	\label{tab:result3}
	\resizebox{\columnwidth}{!}{%
		\begin{tabular}{|l|l|c|c|c|c|c|}
			\hline
			\textbf{Hedef   Eğitim} & \textbf{Kaynak} & \textbf{Loveda Kırsal} & \textbf{Loveda Şehir} & \multicolumn{1}{l|}{\textbf{HWLC16}} & \multicolumn{1}{l|}{\textbf{HWLC18}} & \multicolumn{1}{l|}{\textbf{Mass. Yol}} \\ \hline
			Loveda Kırsal           & -               & 58.8                   & 11.8                  & 24.2                                     & 3.8                                     &                                         \\ \hline
			Loveda Kırsal           & HWLC16            & 58.6                   & 37.2                  & 40.2                                     & 25.9                                    &                                         \\ \hline
			Loveda Kırsal           & HWLC18            & 55.0                   & 39.1                  & 21.2                                     & 28.6                                    &                                         \\ \hline
			Loveda Kırsal           & HWLC1618          & 60.2                   & 38.3                  & 38.6                                     & 25.7                                    &                                         \\ \hline
			Loveda Şehir            & -               & 30.18                  & 62.07                 & 8.2                                      & 15.1                                    &                                         \\ \hline
			Loveda Şehir            & HWLC18            & 36.8                   & 66.35                 & 12.2                                     & 17.7                                    &                                         \\ \hline
			Loveda Şehir            & HWLC1618          & 38.3                   & 60.2                  & 32                                       & 18.6                                    &                                         \\ \hline
			Mass. Yol               & -               & 14.5                   & 0.07                  & 4.43                                     & 19.2                                    & 50.17                                   \\ \hline
			Mass. Yol               & HWLC18            & 19.9                   & 0.06                  & 25.1                                     & 31.7                                    & 58.52                                   \\ \hline
			Mass. Yol               & HWLC1618          & 17.2                   & 0.03                  & 23.9                                     & 31.5                                    & 59.68                                   \\ \hline
		\end{tabular}%
	}
\end{table}

Tablo'da görüldüğü üzere, Loveda Kırsal Veritabanında en yüksek başarım, kaynak küme olarak HWLC1618 kullanıldığında \% 60.2 ile elde edilmiştir. Loveda şehir veritabanıda eğitilen modellerde HWLC18 veri kümesini kaynak küme olarak alındığında \%66.35 başarım elde ederken daha büyük HWLC1618 verikümesiyle alan adaptasyonu, IoU skorunu \% 60.2'ye düşürerek yaklaşık \%6 negatif transfere sebep olmuştur. Massachusets Yol veritabanı üzerinde en yüksek IoU skoru da yine HWLC1618 veritabanından transfer işlemi gerçekleştiğinde yüksek başarımla sonuçlanmıştır.


\section{Vargılar}

Bu makale, uydu görüntülerinden otomatik yol bulma problemi için sözde etiketleme ve en az sınıf karışımı yöntemi tabanlı bir yarı gözetimli bölütleme yöntemine ait ön sonuçlarımızı sunmaktadır. En az sınıf karışımı tabanlı yarı-gözetimli öğrenme yöntemini kaynak küme üzerinde yapılan tahminlerin kararlılığını arttırarak tanıma başarısını arttırmakta kullandık. Ancak daha önce hiç görülmemiş ve karar sınırında olan bölgelerde kötüleşmeler yaşandığı için bu yöntemi sözde etiketleme yöntemiyle üretilmiş etiketlerle destekledik. Bu şekilde aldığımız tahminlerle değişik bölgelere ait ama benzer içerikli veritabanları üzerinde yüksek başarım elde edilebilmiştir.


Kullanılan mimarilerden UNet++ ve HRNet benzer olarak cozunurluk arttıkça daha yüksek IoU oranları elde edilmesine olanak sağlamıştır. İşaret kalitesinin daha yüksek olduğu şehir veri kümelerinde HRNet daha düşük olduğu kırsal veri kümelerinde ise UNet++ yöntemi daha etkin başarı elde etmiştir. İkinci olarak resimlerin HWLC18 verikümesinde Level18'den Level16'ya dönüştürülerek HWLC1618 kümesine katılması, modellerin genelleşmesi üzerinde büyük katkı sağlamaktadır. Bu şekilde boyutları birbirine uygun hale getirilen resimlerde yapılan tahminlerde, genel olarak diğerlerine kıyasla yüksek başarımlar görülmüştür.




Çalışmanın bundan sonraki kısımlarında, sözde etiketleme yöntemleriyle bir ablasyon çalışması yapılması ve daha büyük sayıda işaretsiz resimin kaynak olarak kullanıldığı deneyler yapılması hedeflenmektedir. Bunun dışında, geliştirilen alan adaptasyonu yönteminin, yol sınıfını içeren birden çok sınıflı yeryüzü kullanım haritaları çıkarma problemlerinde kullanımı ve çıkarılan yolların vektörize edilmesi problemleriyle kullanımı üzerine de çalışmalar yapılması olası araştırma hedefleri olarak eklenebilir.

\bibliographystyle{IEEEtran}

\end{document}